\documentclass[10pt,twocolumn,letterpaper]{article}

\usepackage{cvpr}
\usepackage{times}
\usepackage{epsfig}
\usepackage{graphicx}
\usepackage{amsmath}
\usepackage{amssymb}
\usepackage[style=numeric,backend=biber]{biblatex}
\usepackage{microtype}
\usepackage{graphicx}
\usepackage{subfigure}
\usepackage{booktabs} 
\bibliography{bibliography}
\usepackage{xcolor}

\pdfoutput=1 

\usepackage{hyperref}

\cvprfinalcopy 

\def\cvprPaperID{****} 

\begin{document}

\newcommand*\samethanks[1][\value{footnote}]{\footnotemark[#1]}

\title{\LaTeX\ Author Guidelines for CVPR Proceedings}

\title{Keep Your AI-es on the Road: \\Tackling Distracted Driver Detection with Convolutional Neural Networks and Targeted Data Augmentation}
\author{
  Nikka Mofid\thanks{Equal Contribution, Stanford University}\\
  \texttt{nmofid@stanford.edu}
  \and
  Jasmine Bayrooti\samethanks\\
  \texttt{jbayrooti@stanford.edu}
  \and
    Shreya Ravi\samethanks\\
  \texttt{sravi2@stanford.edu}
}

\maketitle

\begin{abstract}
According to the World Health Organization, distracted driving is one of the leading cause of motor accidents and deaths in the world. In our study, we tackle the problem of distracted driving by aiming to build a robust multi-class classifier to detect and identify different forms of driver inattention using the State Farm Distracted Driving Dataset. We utilize combinations of pretrained image classification models, classical data augmentation, OpenCV based image preprocessing and skin segmentation augmentation approaches. Our best performing model combines several augmentation techniques, including skin segmentation, facial blurring, and classical augmentation techniques. This model achieves an approximately $15\%$ increase in F1 score over the baseline, thus showing the promise in these techniques in enhancing the power of neural networks for the task of distracted driver detection.
\end{abstract}

\section{Introduction}
\label{sec:Introduction}
According to the World Health Organization, distracted driving is one of the leading causes of motor accidents. Every year, 1.35 million people across the world are killed as a result of road traffic and many more are injured \cite{WHO:}. Driver inattention is a major contributor to highway crashes, with the National Highway Traffic Safety Administration (NHTSA) estimating that approximately 25\% of police reported crashes involve some form of driver inattention with distracted driving being one of the leading causes \cite{drivingStats:}. Furthermore, recent studies have shown that drivers engaged in live or phone conversations were less aware of traffic movements around them. Thus, distracted driving often leads to dangerous and sometimes deadly situations, and so finding measures to identify and reduce distracted driving has become a major focus in research.

According to the Center for Disease Control and Prevention, there are three major kinds of distraction: visual, cognitive, and manual \cite{CDC:}. In order to tackle the widespread problem of distracted driving in society, in this study we aim to identify distracted drivers using the State Farm Distracted Driving Dataset by building a model that can identify visual and manual types of distracted driving and classify the form of distraction using pretrained image classification models, data augmentation, and OpenCV based image preprocessing techniques. As it is difficult to visually identify cognitive distraction from single images of drivers, we focus on the other two kinds of distraction.

The input to our algorithm is a $224$ x $224$ x $3$ image. We then use a Convolutional Neural Network (CNN) to output a predicted class characterizing the nature of the driver and their distractedness. The $10$ output classes that our model seeks to identify are: ``safe driving, ``texting right", ``talking on phone right", ``texting left", ``talking on phone left", ``operating radio", ``drinking", ``reaching behind", ``hair and makeup", and ``talking to passenger".

Due to a limited number of drivers included in the dataset, creating a model that can generalize well to new drivers and environments not included in the dataset is one of the major challenges we face. We intend to create a robust and generalizable model through a careful crafting of our experiment with the described methodology.

\section{Related Works}
\label{sec:RelatedWorks}
Recent research on improving CNN models have followed an ablation study approach. For instance, Tong He et al used an ablation study to empirically evaluate various refinements' impacts on the final model accuracy, which enabled them to improve ResNet-50's best validation accuracy from 75.30\% to 79.29\% on the ImageNet dataset \cite{ablation1}. Other recent work on pruning state-of-the-art deep CNNs focuses on optimizing the network structure by removing filters and changing their sizes in order to find good combinations of units to keep in the structure that improve training speed and accuracy \cite{ablation4} \cite{ablation3}. We aim to follow a similar ablation study approach keeping the model constant and focusing instead on changing preprocessing augmentation techniques to provide interpretability for which kinds of techniques improve the classifier's performance. Our ablation approach to this problem sets our work apart from other attempts to solve similar problems. For instance, one study seeks to classify images of drivers into similar categories as ours using a number of different CNN models but using two different datasets, one for night and another for day driving \cite{similar}. Our approach has the advantage of greater transparency in techniques but is likely to perform worse on images of drivers at night due to the nature of our dataset. We also use different metrics, namely F1 score, precision, and recall, to gain a deeper understanding of our model's performance beyond accuracy, which has been typically used.

Currently, there are a number of state-of-the-art deep CNN models which would be promising to use for our base model. We consider several papers, including a study developing a distracted driver detection system, based on images taken from a camera observing the driver. In one such study, the researchers experimented with using pretrained VGG-16, AlexNet, GoogleNet, and ResNet models to identify distracted drivers. They found that the ResNet model converged the fastest of the four models, however had the highest rate of misclassifications \cite{article1}. Because distracted driving can endanger lives, it is critical to investigate all possible cases of distracted driving, so the model should have a low false negative rate but can have a higher false positive rate. Therefore, considering the results of this study, we decided to use the ResNet-50 model architecture as our base model for the ablation study. The ResNet-50 model has also independently been shown to have good performance in classifying distracted driver behaviors in similar studies \cite{article4}.

As mentioned in the \hyperref[sec:Introduction]{\color{blue}Introduction}, one major challenge we face is the limited number of drivers and constant point of view across all images in the dataset. Thus, we are setting out to not only achieve high performance on the test set, which is equally non-representative of real world images as the training set, but also to find promising methods of identifying distracted drivers in more generalized images with different drivers and from slightly different points of view. To address this, we intend to experiment with techniques for improving the generalization of our model. Recent research supports the use of data augmentation for improving model generalization as it exploits domain knowledge to increase the amount of training data and improve the generalization without reducing the effective capacity or introducing model-dependent parameters, in contrast to regularization, which also improves generalization but blindly reduces the effective capacity of the model \cite{dataAug4}. We follow data augmentation procedures laid out by Mikolajczyk et al to compare and analyze multiple methods of augmentation on-the-fly for our task of image classification, including classical image transformations like rotating, cropping, and color jittering \cite{dataAug2}. 

To further combat the challenge of generalizing to different drivers, we need to encourage the model to consider the driver's posture and hand position rather than attending too much to the driver's facial features or skin. Facial blurring via a Gaussian blur of appropriate noise-determined variance has been shown to be an effective method of anonymizing people for privacy or surveillance reasons \cite{faceBlur}. We intend to apply a similar simple procedure to discourage the model from overfitting to the individual drivers in our dataset. Additionally, recent work has shown skin color segmentation to be useful for various applications including posture detection and hand gesture analysis \cite{skin1}. To address the challenges of varying illumination and background clutter that come with skin segmentation, we will use a combination of the chrominance channels of the HSV, YCbCr and Normalized RGB color spaces to achieve better accuracy for skin segmentation than in just single color spaces \cite{skin2}. 


\section{Dataset and Features}

\subsection{Dataset Details}
We are using the Distracted Driving dataset provided by State Farm, which contains images of drivers, distracted and non-distracted, taken by constant-placed 2D dashboard cameras of dimension $640$ x $480$ pixels in RGB \cite{data}. The dataset contains $22,424$ labelled unique images of drivers that fall into one of the following $10$ distracted driver classifications (as described in \hyperref[sec:Introduction]{\color{blue}Introduction}): ``safe driving, ``texting right", ``talking on phone right", ``texting left", ``talking on phone left", ``operating radio", ``drinking", ``reaching behind", ``hair and makeup", and ``talking to passenger". The data used to label these images are in CSV form that lists the driver subject number with their distraction "class" and unique image identification number. The class distribution of the images is mostly even, but two of the classes (``safe driving" and ``reaching behind") vary about $10\%$ from the median number of images per class and one of the classes (``hair and makeup") varies about $20\%$ from the median, which can affect the class predictions that our model prioritizes.

Our problem is technically challenging due to the fact that there are only $26$ unique drivers in the dataset, so we  need to be careful not to overfit to the drivers in our training data. In addition, the images of drivers in different distracted poses were taken from frames of videos collected by State Farm, thus many of the images are similar, which further exacerbates the overfitting problem.

\begin{figure}[h!]
\center
   \includegraphics[width=8cm]{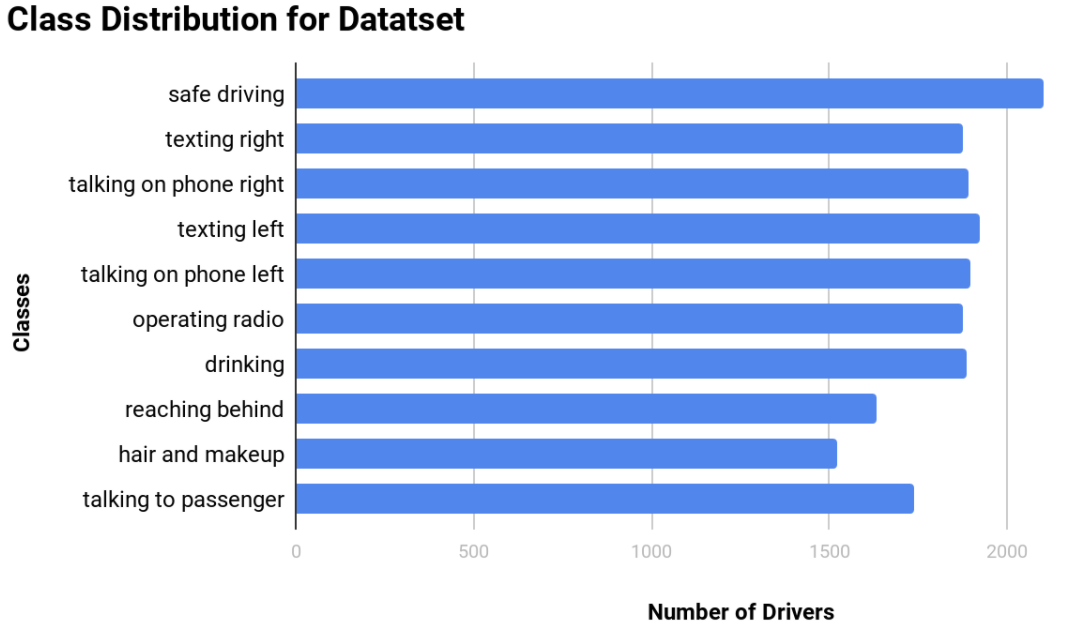}
   \caption{Distribution of classes in the distracted driver dataset}
\end{figure}

\begin{figure}[h!]
  \centering
    \includegraphics[width=4cm]{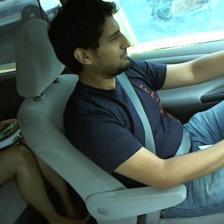}
    \caption{Example Image of Distracted Driver from State Farm Distracted Driving Dataset}
    \end{figure}

\subsection{Data Preprocessing}
\label{sec:DataPreprocessing}
For our baseline model, we utilized ResNet-50 with pretrained weights. In order to enhance model performance,  we used preprocessing steps that resized the image to $256 \times 256$ spatial dimensions, center cropped to $224 \times 224$ spatial dimensions normalized, and normalized with a fixed mean and standard deviation (specifically a mean of $[0.485, 0.456, 0.406]$ and a standard deviation of $[0.229, 0.224, 0.225]$). We let our ResNet-50 baseline model perform the feature extraction.

As part of our experimentation, we performed several other preprocessing and image processing steps on our train and test data in order to prevent the model from overfitting on certain features of the driver (i.e the drivers face) and help the model focus on the important parts of the image (i.e the driver's posture). As described in the following sections, we used OpenCV to identify an eye in the image and blurred out a region relative to the eye that approximately covered the face of the driver in order to combat overfitting to the driver's facial features and also performed skin segmentation, highlighting the driver's skin while blacking out the rest of the image in order to draw the models attention to the driver's posture. Afterwards, we ensembled our methods, including classical data augmentation techniques, to see if we could use them in conjunction build a strong classifier.

\subsection{Train and Test Split}
In our initial approach, we decided to split our data into $80\%$ train data and $20\%$ test data, resulting in approximately $18,000$ train images and approximately $4,500$ test images. When we ran our pre-trained ResNet-50 model with additional fully connected layers for our 10-class classification problem, we got results that seemed too good, and we realized that this was likely because our train and test set were incredibly similar due to the aforementioned limited number of drivers and limitations of video data being treated as individual images. Because of this, the model was overfitting on both the test and train data (despite not seeing the test data during train time) but would likely not generalize well to any new drivers and environments. Thus, we decided to take a new approach with the the train and test split.

We chose $5$ drivers at random from the $26$ (because this is approximately $20\%$ of the data) and included all of their images in the test data. We used data from the other $21$ drivers for the training set, thus ensuring minimal overlap between the two datasets. Using this approach, our baseline results were more realistic and would likely generalize well to new drivers.

\section{Methods}
\label{sec:Methods}


\subsection{Baseline}

ResNet has been a state-of-the-art convolutional model since it won the ImageNet classification challenge in 2015. This model presents residual modules as a solution to the problem of optimizing deeper networks, thereby enabling researchers to take advantage of the greater representation power of deep networks while not trading off on training error. These residual modules learn a residual mapping rather than directly trying to fit the desired mapping underlying the data.

\begin{figure}[h!]
\center
\includegraphics[width=4cm]{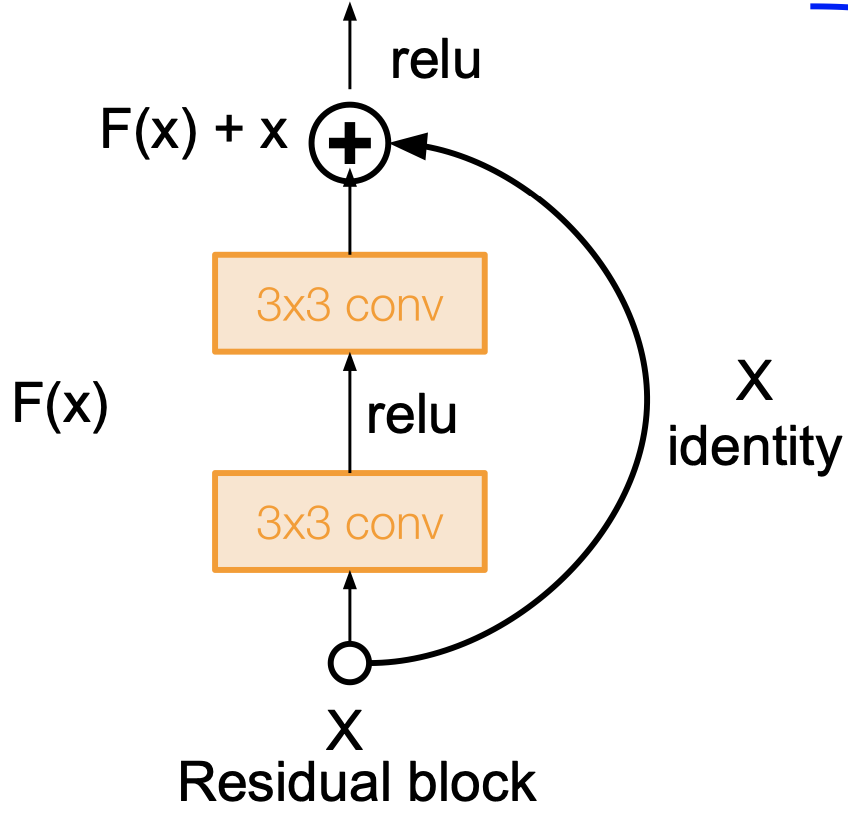}
\caption{Residual Module Architecture}
\end{figure}

The ResNet model includes stacks of residual blocks, consisting of two $3$ x $3$ convolutional layers to learn the residual mapping, as shown in figure 3. The full architecture is shown in the appendix. As described in the \hyperref[sec:RelatedWorks]{\color{blue}Related Works} section, we chose to use ResNet-50 for our model as it performed well on similar challenges and converged quickly. Thus, we are using a pretrained ResNet-50 as our baseline model with a fully connected layer of hidden size $512$, a ReLU nonlinearity layer, a dropout layer (with probability $0.2$), and two fully connected layers of output size $10$ (for the $10$ distracted driver classes). For our baseline, we trained these last layers of the model using our driver-separated training set with the ResNet-50 specific preprocessing detailed in \hyperref[sec:DataPreprocessing]{\color{blue}Data Preprocessing}. We trained the model for 25 epochs using an Adam optimizer and achieved decent results, as specified later. For the loss function, we use cross-entropy loss with multi-class softmax where M is the number of classes ($10$ for our problem):

$$Loss = -{(y\log(p) + (1 - y)\log(1 - p))}$$
$$-\sum_{c=1}^My_{o,c}\log(p_{o,c})$$


\subsection{Classical Data Augmentation}
\label{sec:DataAugmentation}

Data augmentation is a popular technique for improving model generalization and preventing overfitting by allowing the validation error to continue decreasing with the training error. This is because augmented data represents a more comprehensive input space, through which the differences between the training and any possible testing sets are more likely to be minimized \cite{dataAug}. Since overfitting and generalization are the two main challenges that we face with our dataset, we decided to experiment with two on-the-fly data augmentation techniques---specifically, random rotation and brightness adjustment.

\begin{figure}[h!]
\subfigure[Image of Driver with Random Brightness Adjustment]{\includegraphics[width=4cm]{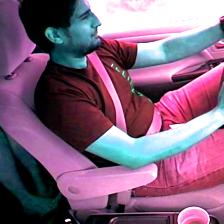}}
\subfigure[Image of Driver with Random Rotation Adjustment]{\includegraphics[width=4cm]{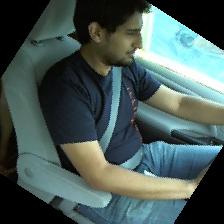}}
\caption{Image Preprocessing and Augmentation Examples}
\end{figure}

Since all the drivers in the dataset were facing to the right, we used random rotation augmentation to produce examples of our driver rotated slightly between $0$ and $45$ degrees in both directions. In order to further increase our dataset size, we decided to add a second augmentation introducing a small, random amount of color jitter (characterized by brightness, saturation, contrast, and hue adjustments) to add a non-spatial based form of noise into our dataset. It is important to note that we do this data augmentation ``online'' or on the fly, performing the transforms using the torch transforms library each time the train and test data is loaded into our model. 

The input to our random rotation and brightness adjustment transforms is the images from the State Farm Distracted Driving Dataset, resized and cropped to fit the ResNet-50 standard size of $224$ x $224$ x $3$ and with the appropriate color normalization, as explained in \ref{sec:DataPreprocessing}. We use the Pytorch transforms library in order to perform this random rotation and color jitter. The output of this procedure is two sets of the transformed input images. We then concatenated these transformed image sets with our original input images and shuffled, giving an augmented set of three times the size of our original training dataset. Numerically, using these two augmentation techniques in tandem, we produced a training set of $55,032$ images from the original training set of $18,344$ images.

\subsection{OpenCV Facial Blurring}
\label{sec:BlurredImages}

In order to further combat the challenge of overfitting to the drivers and generalizing to different kinds of images, we need to encourage the model to consider the driver's posture so that it does not memorize patterns in the appearance of the 21 drivers' faces. As described by T. Muraki et al, facial blurring via a Gaussian blur is an effective way of anonymizing people for privacy purposes \cite{faceBlur}. The same technique, called Random Blur, can be applied to our problem to blur our the driver's face in order to discourage the model from overfitting to the individual drivers in our dataset. We took inspiration from the random blur techniques and experimented with facial blurring as an augmentation technique on both the training and test sets.
 
In order to perform facial blurring, for each train and test image, we identified the position of the driver's eyes via a pretrained eye detector from haarcodes in OpenCV \cite{haar}. From the eye size and location, we then extrapolated the coordinates to find the approximate area corresponding to the driver's face and blurred out the facial region by smoothing over the image in that approximate area using a Gaussian Blur with appropriate variance. We then used these images in combination with the original input images to assemble new Facial Blur augmented training and test sets for a ResNet-50 model with pretrained weights.

\begin{figure}[h!]
\subfigure[Original Image of Driver]{\includegraphics[width=4cm]{figure2.jpg}}
\subfigure[Image of Driver with OpenCV Facial Blurring]{\includegraphics[width=4cm]{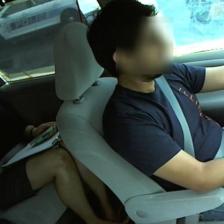}}
\caption{Facial Blurring Example}
\end{figure}

The inputs to our facial blurring script were the images from the State Farm distracted driving dataset resized and cropped to fit the Resnet-50 standards as explained in
\hyperref[sec:DataPreprocessing]{\color{blue}Data Preprocessing}. The outputs of this blurring script are all the same images from the train and test sets with the drivers' faces blurred. Unlike the classical augmentation method, we decided to perform this technique offline since it is is more complex and time consuming. Using this technique, we were able to double the size of our training set while improving the generalization of our model.

\subsection{Skin Segmentation}
\label{sec:skinseg}
After performing our first set of experiments and analyzing the saliency maps from our models, we noticed that the model was focusing on the posture, particularly arm and hand position of the driver, for many of the successfully classified images. Thus, we hypothesized that if we could encourage the model to focus more on the driver's posture we would be able to improve the performance of our classifier and help the model generalize better to new images of drivers.

Skin segmentation is a technique that has been widely used in many biometric applications, face recognition, and hand gesture recognition. It is the act of separating skin and non-skin pixels in an arbitrary image and has been shown to be effective for posture detection and hand gesture analysis, as discussed in \hyperref[sec:RelatedWorks]{\color{blue}Related Works}. There are many different methods of performing skin segmentation. In this study, we focus on the method presented by R. Rahmat et al, which combines the lighting channels of the HSV, YCbCr and Normalized RGB color spaces to obtain high accuracy of skin detection \cite{skin2}. This method enables us to find a simple, effective, and efficient way to identify the skin of a driver and black out the rest of the image to teach our model to focus on a driver posture. 

We implemented a version of this combination of ``chrominance'' technique based on Will Brennan's Skin Detector, \cite{Brennan} which uses masking and thresholding of RGB (red, green, blue values) and HSV (hue, saturation, value) through the OpenCV library in order to leave the colored the portion of the image which contains skin (falling within the range of RGB and HSV defined by the user) intact and blacks out the rest of the image. In order to detect the range of colors which we would consider ``skin'', we wrote a python script to allow us to toggle the RGB and HSV values of our input images to find a range of RGB and HSV values that would, on most images, black out the background and leave the skin intact. It is important to note that there is some bias in this technique as it does not work as well on persons of color and is dependent on the lighting in the image, but as there is little ethnic diversity in our dataset and the images are lighted homogeneously, we were able to find an RGB and HSV color range that fit many images.

The inputs for our skin segmentation script were the original set of images from the distracted driving dataset and the ranges of RGB and HSV values, which we would use as the ranges for skin segmentation. The outputs are used as input images to the model and show the original image with only the skin of the driver remaining in color and the background blacked out. As with the facial blurring set, we produced these images offline as the generation was complex and time consuming. We then used these images to augment our training set by resizing and cropping them to fit the ResNet-50 standard size of $224$ x $224$ x $3$ and the appropriate color normalization, and then concatenating them with our original input images and shuffling. Through this technique we were able to double the size of our dataset while also teaching our model to focus on the driver's posture and less on any background noise.

\begin{figure}[h!]
\subfigure[Original Image of Driver]{\includegraphics[width=3.5cm]{figure2.jpg}}
\subfigure[Image of Driver with OpenCV Facial Blurring]{\includegraphics[width=4.5cm]{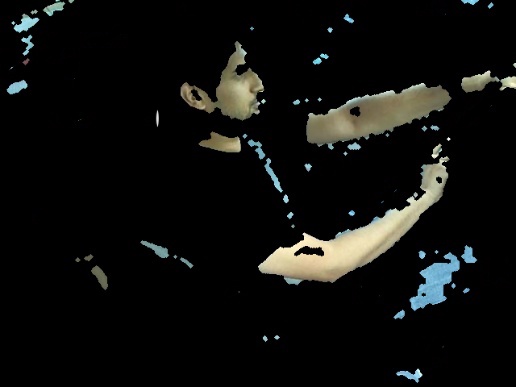}}
\caption{Skin Segmentation Example}
\label{skinsegexample}
\end{figure}

\subsection{Combining Preprocessing Augmentation \\ Techniques}
\label{sec:combined}
Finally, through our experimentation, we found that we had developed several different powerful pre-processing augmentation techniques. In order to round out our ablation study, we decided to explore combining these techniques to see if a specific combination of pre-processing augmentation steps could boost our model's performance. As such, we experimented with different combinations of the techniques by concatenating together the appropriate datasets we had generated for facial blurring and/or skin segmentation and generating the classically augmented images on-the-fly. We found that by combining a number of techniques, we were able to achieve high scores on our multi-class classification task, as described further in the \hyperref[sec:Results]{\color{blue}Results and Discussion} section.

\section{Results and Discussion}
\label{sec:Results}

\subsection{Hyperparameters}
We tuned a few main hyperparameters such as learning rate, mini-batch size, and probability of dropout in the final dropout layer (introduced as part of the final classification layers of the model). We started the learning rate and betas of the Adam optimizer with the default recommended values ($10^{-3}$ for the learning rate and $[0.9, 0.99]$ for the betas), and tuned the learning rate to be $3*10^{-3}$ because it allowed for faster training with slightly improved results. We used a mini-batch size of $64$ images to fit the memory requirements of our CPU and GPU and allow for faster convergence while also being large enough to have minimal noise in the calculated gradients. Finally, we chose the dropout probability to be 20\% (meaning there is an 80\% chance of retaining a given unit) based on the recommendations of the paper that introduced dropout as a regularization technique \cite{dropout}.

\subsection{Evaluation Metrics}
We used F1 score to evaluate our best model as well as accuracy, precision, and recall as metrics to gain a quick understanding of where our model tended to fail, which then informed the further quantitative and qualitative metrics we examined.

F1 score is a measure of model performance that gives a summary of the precision and recall metrics and is calculated as follows:

\[\texttt{F1} = 2 \times \frac{\texttt{precision} \times \texttt{recall}}{\texttt{precision} + \texttt{recall}}\]

Precision gives a measure of how many of the images predicted to be in a given class $c$ actually belong to that class, and is calculated as follows:

\[ \frac{\sum_{c \in C} \texttt{true\_positives}_c}{\sum_{c \in C} \texttt{true\_positives}_c + \texttt{false\_positives}_c}\].

Recall gives a measure of how many of the images that belong to a given class $c$ are actually predicted to be in that class, and is calculated as follows:

\[\frac{\sum_{c \in C} \texttt{true\_positives}_c}{\sum_{c \in C} \texttt{true\_positives}_c + \texttt{false\_negatives}_c}\].

\subsection{Quantitative Results}
\begin{figure}[h!]
   \includegraphics[width=8.5cm]{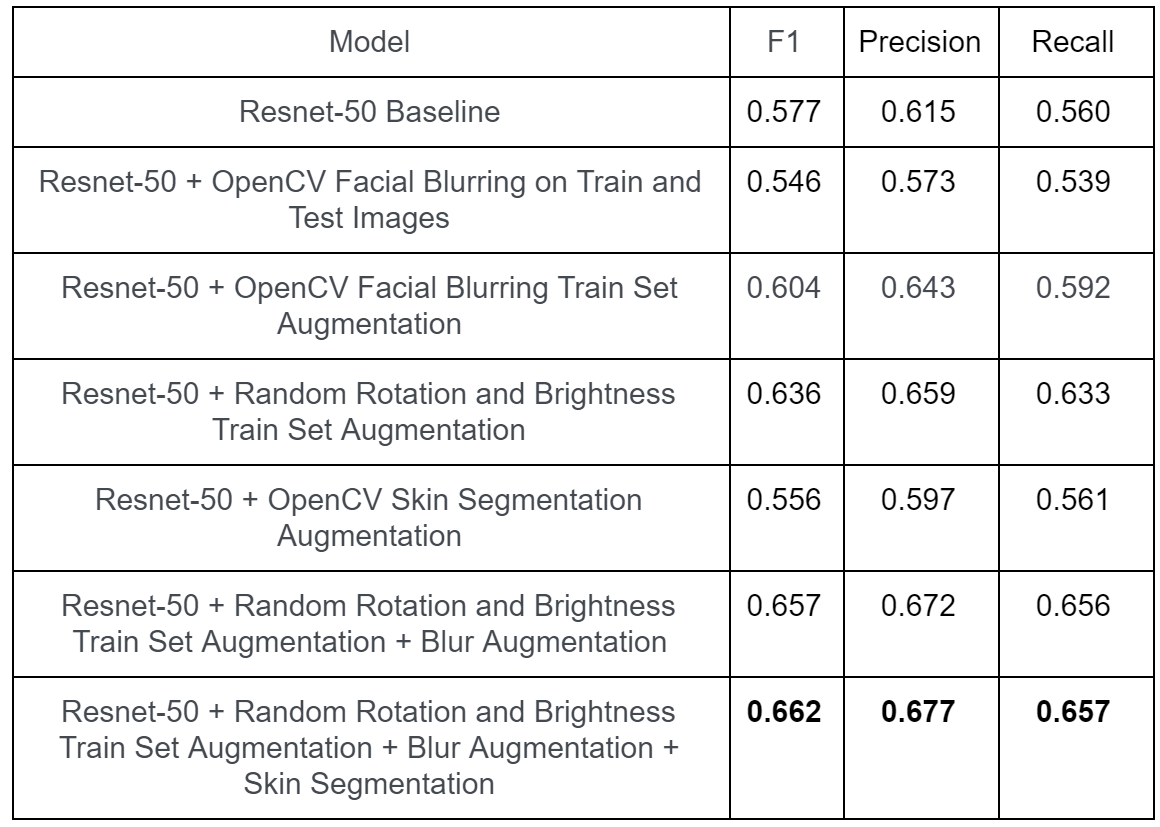}
    \caption{Experiment Results}
   \label{resultstable}
\end{figure}

\begin{figure}[h!]
\subfigure[Confusion Matrix for ResNet-50 Baseline]{\includegraphics[width=4cm]{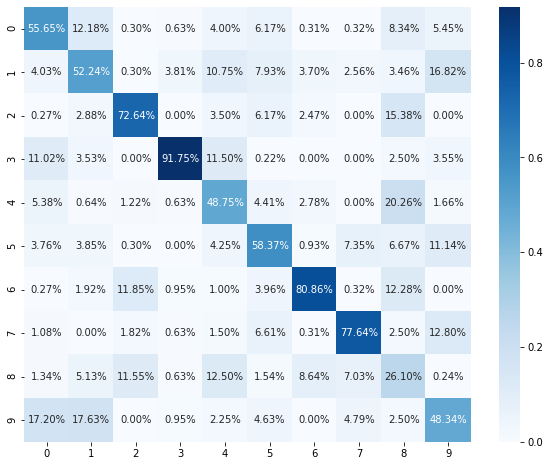}}
\hfill
\subfigure[Confusion Matrix for ResNet-50 with Facial Blurring, Skin Segmentation, and Classical Augmentation]{\includegraphics[width=4cm]{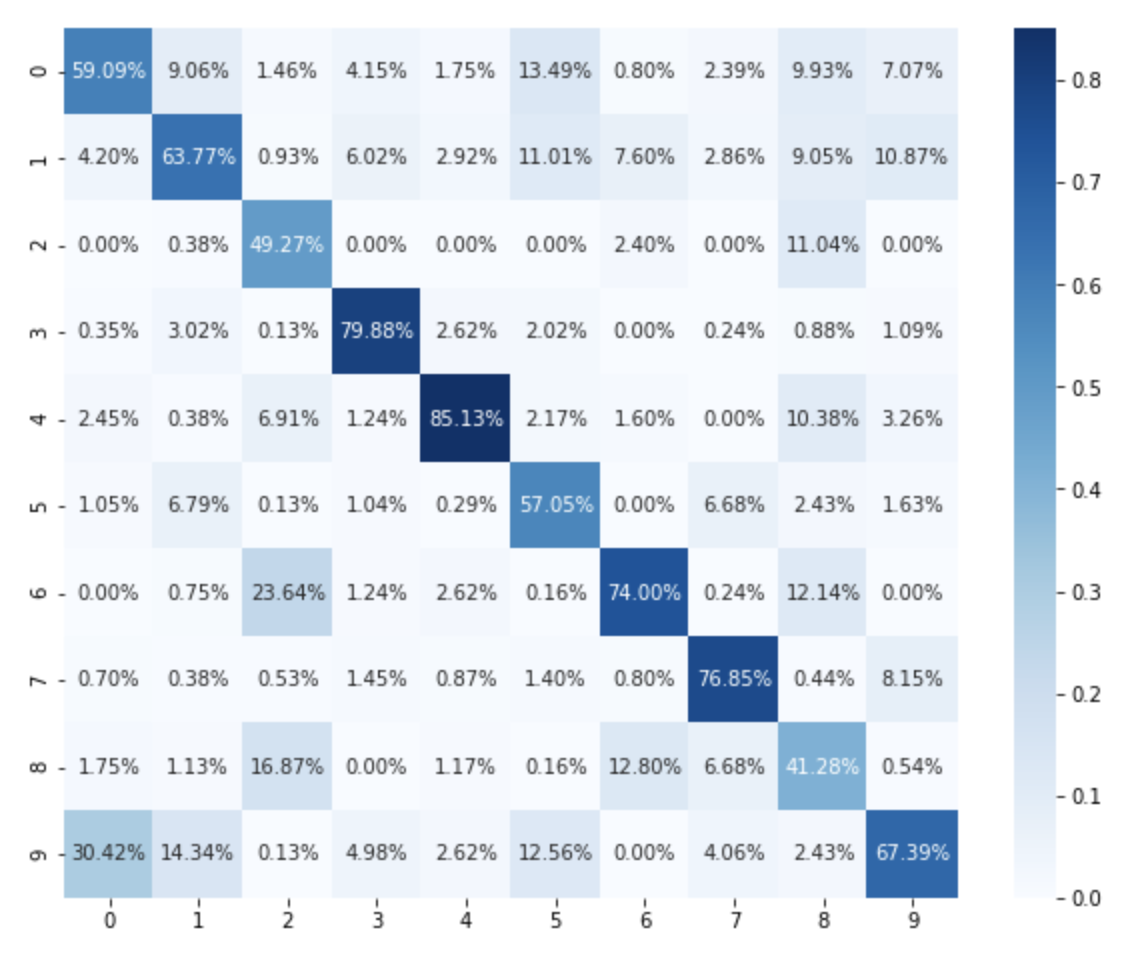}}
\caption{Experiment Confusion Matrices}
 \label{confusion}
\end{figure}

The results of our experiments are detailed in the table in Figure \ref{resultstable}. We also generated the Confusion Matrices in Figure \ref{confusion} and the Appendix to provide more insight into our model's behavior. Please refer to the \hyperref[sec:Methods]{\color{blue}Methods} section for further explanation of our experiments.

\subsection{Qualitative Results}
The Saliency maps for our different model are shown below in Figure \ref{qualitative}.

\begin{figure}[h!]
   \includegraphics[width=8cm]{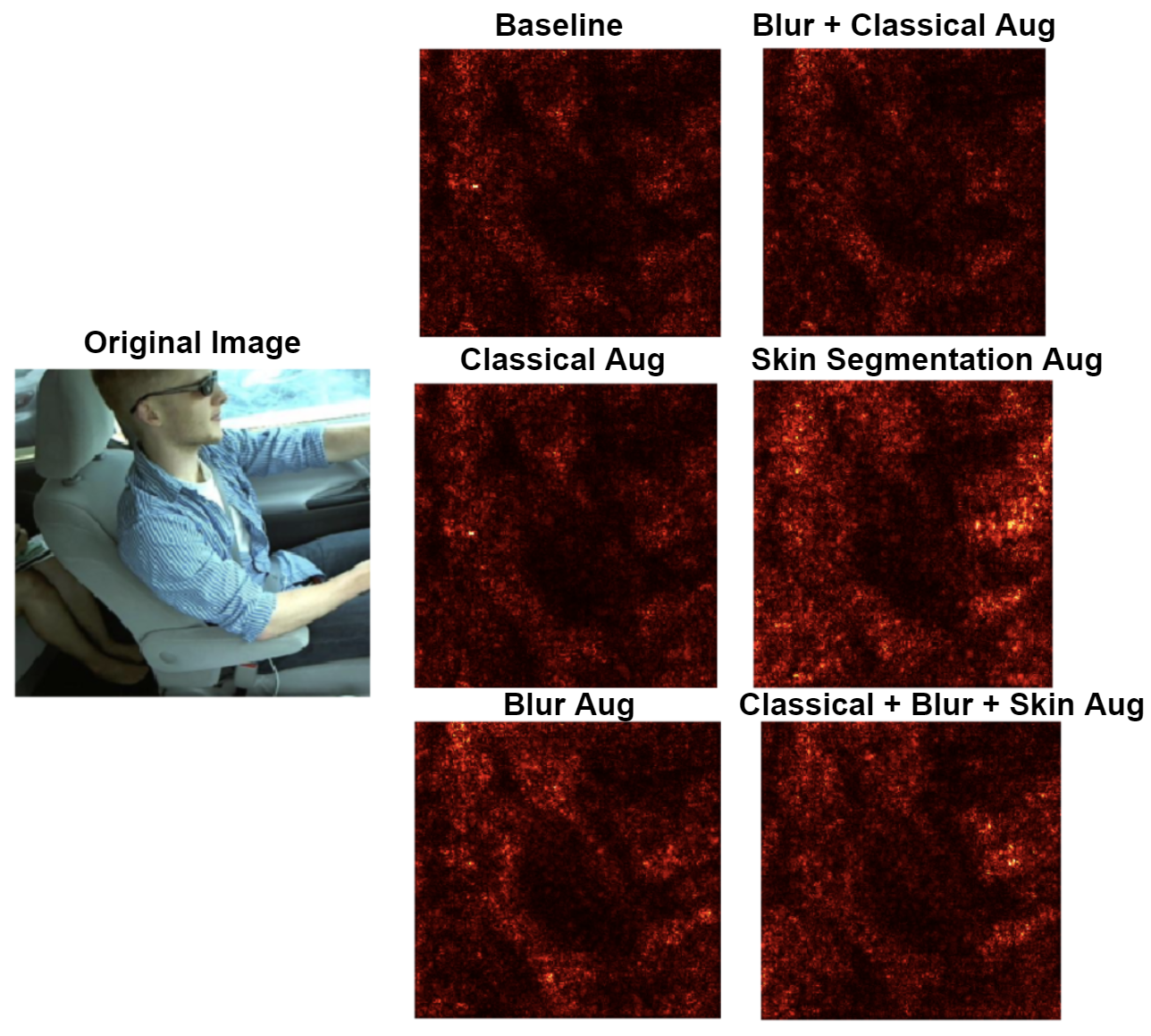}
    \caption{Saliency Maps for our Experiments}
   \label{qualitative}
\end{figure}

\subsection{Quantitative Results Discussion and Analysis}
\subsubsection{Results Table Analysis}
Looking at Figure \ref{resultstable}, one can see a breakdown of our ablation study to build a strong classifier for Distracted Driving. It is clear that adding pre-processing data augmentation techniques had a significant positive impact on performance.

First of all, we found that using data augmentation significantly improved the performance of the ResNet-50 model. This is likely because data augmentation builds up the model's tolerance to noise so it can better generalize to new images in the test set. This is particularly important for addressing our dataset limitations of having very few drivers present in our overall dataset and discouraging the model from memorizing patterns specific to these individuals. We observed a considerable increase in performance (${\sim}10\%$ as a result of augmenting the data using the techniques described in \hyperref[sec:DataAugmentation]{\color{blue}Classical Data Augmentation}. This is likely because we used augmentations that most similarly represented data that the model would likely encounter, such as slight rotations and slight color changes.

For our next experiment, we directly blurred out the driver's face to prevent the model from learning about facial specific features. We believed this could improve our model's performance over the baseline by teaching the model to focus more on the posture of the driver rather than the face, a more salient feature. However, we found that this actually led our performance to decrease, leading us to conclude that the drivers face's contains important information that the model uses to make its predictions, perhaps by tracking eye placement. So instead, we used facial blurring as an augmentation technique so the model would learn to treat facial features as noise. This approach gave an improvement on the baseline and seems promising as an augmentation technique in conjunction with rotation and color jitter. We hypothesize this is because it allows the model to learn the key facial features necessary while also learning to focus on posture as well.

Seeing improvement in our results by augmenting with blurred train data, we decided to explore augmenting with skin segmented data as detailed in \hyperref[sec:skinseg]{\color{blue}Skin Segmentation}. The model performed slightly worse than the baseline when augmented with only skin segmented data, but achieved the best performance when augmented with skin segmented data in tandem with our classically augmented and facial blurring dataset. We think that, with skin segmentation augmented images alone, the model performed slightly worse because it needed some information from the background, particularly the steering wheel, which we fully blacked out and that skin segmentation augmented images alone did not help solve the overfitting problem on the train set.

Thus, we conclude that our final model, which combines all the image pre-processing methods, performs best because together the images help our model optimally combat the overfitting problem while still providing enough information about salient parts of the image to optimize its classifications. This model also likely performs so well because it receives the most data as the input is augmented by classical augmentation techniques of random rotation and brightness as well as blur augmentation and skin segmentation augmentation, leading the input data to be four times as large as the original training set size.

\subsubsection{Confusion Matrix Analysis}

A confusion matrix shows the summary of predictions for a classification problem with the x axis showing the model's class predictions and the y axis showing the true class labels. Each element $(a,b)$ in the confusion matrix shows the probability of the true class being $b$ given the model predicted the class to be $a$. In Figure \ref{confusion} we show the confusion matrices for a few of our different experiment models.

The expectation of the probability for a random model to predict a given class correctly is $10\%$. The confusion matrix for the ResNet-50 baseline in Figure \ref{confusion} illustrates that our baseline model performed significantly better, predicting any given class correctly with a probability of over $45\%$ for $9$ out of $10$ classes. We did notice however some concerning results: that the probability of correctly classifying a class 8 image, ``hair and makeup", was only 26.1\%, with these images being misclassified as class 4 and class 2 images with 20.26\% and 15.38\% probabilities respectively. Although classes 2 and 4 are likely to be confused with each other because they both represent drivers talking on the phone, it is concerning that the model confuses these classes with ``hair and makeup". We address this problem using data augmentation because class 8 had the least training data in the original dataset.

We observe that the class probabilities of correct classification improve as a whole when using the full ensemble model, consisting of facial blurring and skin segmentation images in addition to classically augmented images. For instance, the probability of correctly classifying class 9 images, ``talking to passenger", increases from 48.34\% in the baseline to 67.39\% in the ensembled model. Similarly, we see that adding more positive training samples for class 8, which had the least amount in our original dataset, improved its classification accuracy from 26.10\% in the baseline to 41.28\% in the ensembled model. Interestingly however, the probability for correctly classifying class 3 images drops from 91.75\% in the baseline model to 79.88\% in the ensembled model. Although this may seem like a drop in performance, the original high accuracy was likely due to the model not predicting images to be class 3 very often, which is why the probabilities in the class 3 row other than the the correct class are high for the baseline model, but are low for the ensembled model. This means that the ensembled model still did better in differentiating between the classes and more often predicted class 3, meaning our ensembling steps have helped increase the robustness and generalization of our model.

We also observed the trend that, across most of our confusion matrices, safe driving has a fairly low false positive rate except for class 1 images in some cases. This means that identifying distracted driving has a low false negative rate but a higher false positive rate, as we had hoped in the \hyperref[sec:RelatedWorks]{\color{blue}Related Works} section.

\subsection{Qualitative Results Discussion and Analysis}
We generated the saliency maps shown in Figure \ref{qualitative} in order to see what parts of the image the model attended to, how that may have affected the model's behavior, and how we could address the issues with our different experiments. Considering the saliency map of an image correctly classified as safe driving, we observe that the model was primarily focusing on the arms and posture of the driver.

Seeing that this was a successful signal for the model, we decided to amplify it, at first by blurring the faces of the driver and adding those images as a supplement to the training images in order to teach the model to treat the faces as noise and not overfit to them so that the model would focus more on posture of the driver when identifying distraction. Considering the saliency map for the output of the blur augmented images model, we see a higher concentration of red on the arms of model and key chest and neck points, which shows that the model focused more on the body of the driver as desired. This is likely one the reasons that the model which used the blur augmented images performed even better than the baseline.

Along the same line of reasoning, we decided to augment the input images to the model with skin segmented images, where only the drivers' posture was highlighted and the rest of the image was blacked out as shown in Figure \ref{skinsegexample}. As shown in the results table, we actually found that augmenting with skin segmented images alone led the model to do slightly worse than the baseline. Looking at the saliency map, this is likely because the model was trying to make sense of the blacked out portions of the images while also focusing more on the arms and posture of the driver. Thus, the model may have overfit to these blacked out regions, which we had intended to be ignored, thereby causing the model to perform worse than the baseline.

Therefore, the model did the best using a combination of all the different pre-processing augmentation techniques since the saliency maps show that the model focused very clearly on the driver's posture while ignoring the background and paying less attention to the face. In turn, the model did not overfit to the train images, which we hypothesize is because the classically augmented images provided the model with more of a variety to train on. Looking at the saliency map, it is clear to see that with a combination of all our data pre-processing augmentation techniques, the model learns to focus on the most salient features of the image and ignore noise, thus leading it to perform very well.  

\section{Conclusion and Future Work}
As shown in \hyperref[sec:Results]{\color{blue}Results and Discussion}, we observed significant improvement over our baseline when augmenting the dataset with random rotation and brightness adjustment, facial blurring, and a combination of all augmentation methods in addition to skin segmented images. The final method described achieved the highest scores with an F1 score of $0.662$. This combination of all pre-processing augmentation methods performed the best as the augmented images highlighted different key features which helped combat the overfitting problem and enabled the model learn to generalize well while focusing on the most important features for classifying distracted driving. Additionally, using all these pre-processing augmentation steps quadrupled the original training set size, which helped boost performance by providing more learning examples. Our study shows that through thoughtful image processing techniques and careful analysis of model behavior, one can teach a model to focus on important characteristics salient to the given problem while reducing overfitting, hence leading to a more robust model.

For future work, we are interested in experimenting with changes to our model's architecture such as adding in Spatial Transformers to allow spatial manipulation of data within the network and increase the model's spatial invariance to the input data. Spatial transformer networks have been shown to be effective for mitigating reliance on pre-set data augmentation techniques while improving model generalizability and have been integrated into standard models like PointNet \cite{pointNet}. We hope that inserting the spatial transformer modules into the ResNet-50 architecture will similarly improve our image classifier by enabling the model to learn invariance to scale, rotation, croppings, and non-rigid deformations. We are also interested in exploring unfreezing layers of our model, experimenting with more powerful ResNets and other pretrained models, and ensembling different pretrained models in combination with our preprocessing data augmentation techniques to try to further boost performance.

We are very excited for future possibilities building on the work we have laid out in this study!

\newpage

\section{Appendix}

\begin{figure}[h!]
\center
\includegraphics[width=6cm]{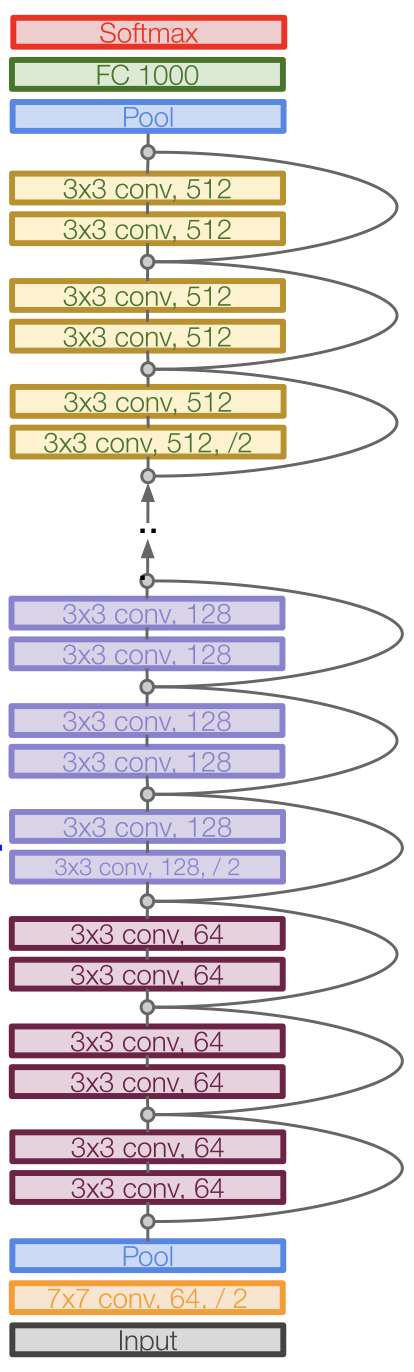}
\caption{ResNet Architecture}
\end{figure}

\begin{figure}[h!]
\subfigure[Confusion Matrix for the Resnet-50 Baseline]{\includegraphics[width=4cm]{figure9.png}}
\subfigure[Confusion Matrix for Resnet-50 with Classical Augmentation]{\includegraphics[width=4cm]{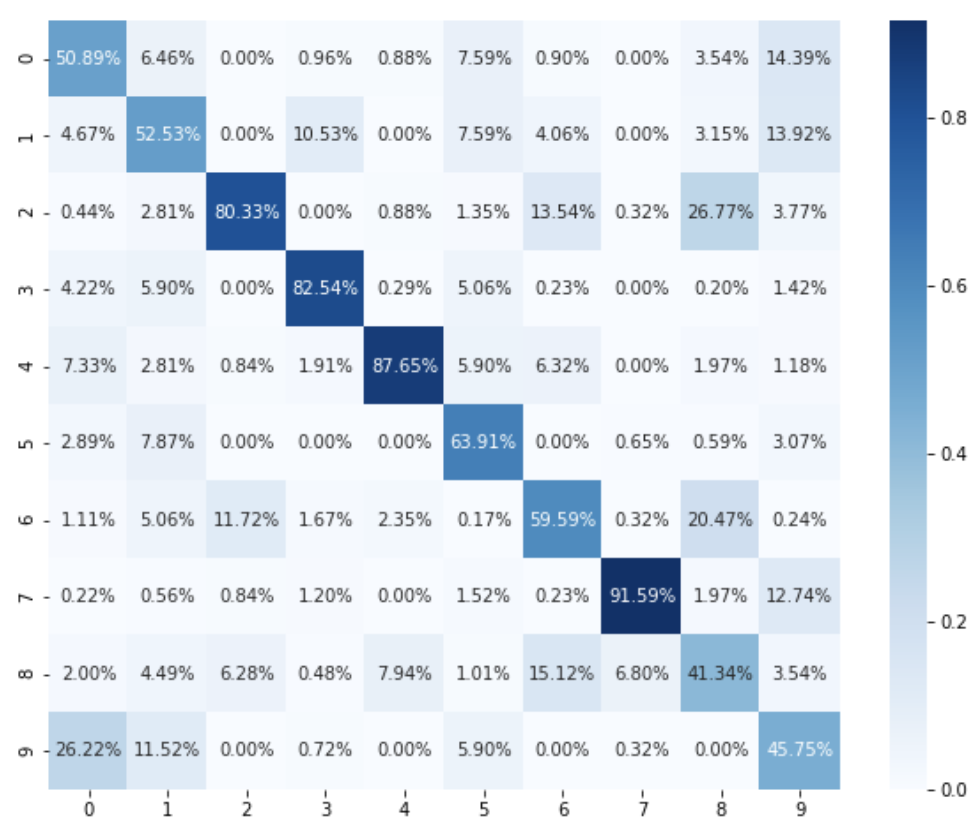}}
\subfigure[Confusion Matrix for Resnet-50 with OpenCV Facial Blurred Augmentation]{\includegraphics[width=4cm]{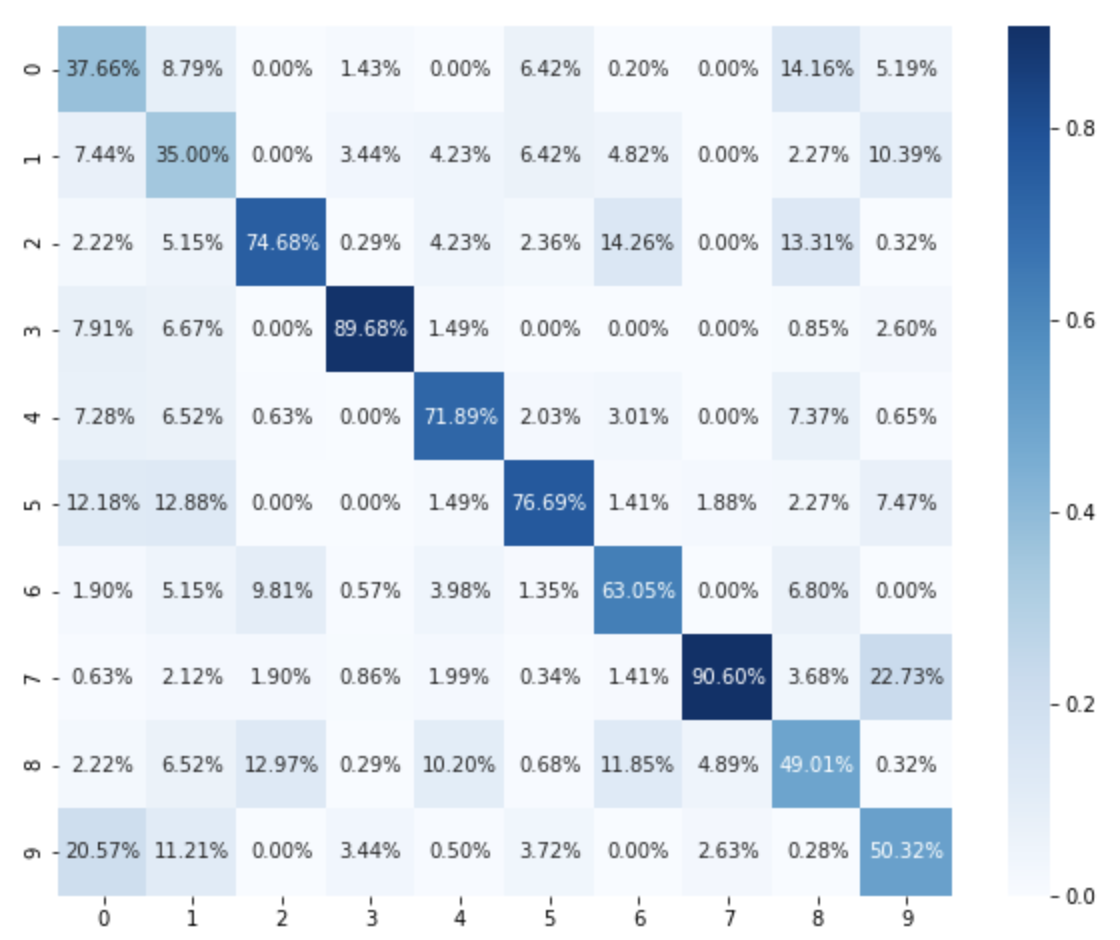}}
\subfigure[Confusion Matrix for Resnet-50 with OpenCV Skin Segmentation Augmentation]{\includegraphics[width=4cm]{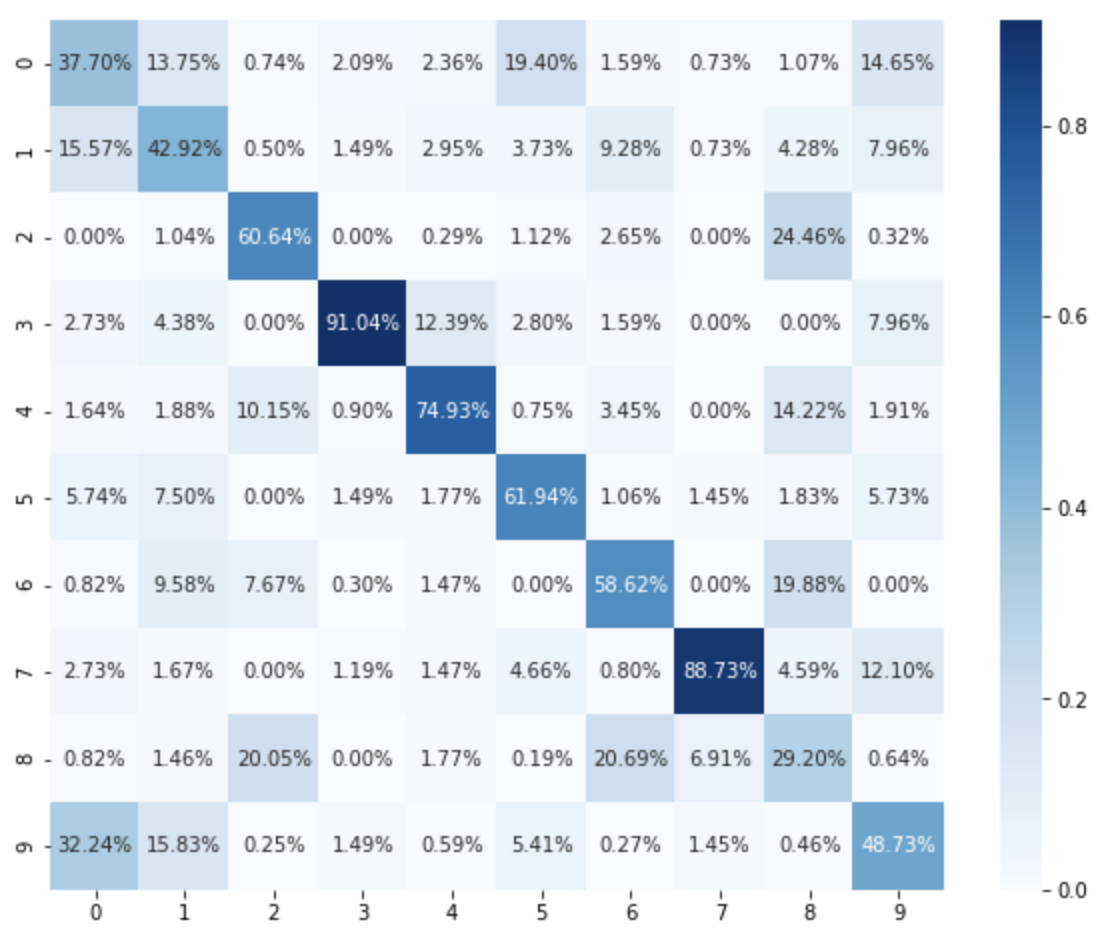}}
\subfigure[Confusion Matrix for Resnet-50 with Full Ensemble Augmentation]{\includegraphics[width=4cm]{figure10.png}}
\caption{Experiment Confusion Matrices}
\end{figure}

\break
\twocolumn

\printbibliography

\end{document}